\documentclass{article} %
\usepackage{iclr2027_conference,times}

\PassOptionsToPackage{hyphens}{url}
\usepackage[hidelinks]{hyperref}
\usepackage{url}
\usepackage{graphicx}
\usepackage{booktabs}
\usepackage{multirow}
\usepackage{tabularx}
\usepackage{amsmath}
\usepackage{amssymb}
\usepackage{mathtools}

\DeclareMathOperator*{\argmax}{arg\,max}

\title{Simple Meta-Learning for HPO and CASH \\ Using Performance Representations}

\author{Emiel Witting, Cheng Yan \& Tom Julian Viering \\
Delft University of Technology \\
Delft, The Netherlands \\
\texttt{emiel.witting@gmail.com, \{C.Yan-1,t.j.viering\}@tudelft.nl}}

\iclrfinalcopy

\begin{document}

\maketitle
\lhead{}
\renewcommand{\headrulewidth}{0pt}

\begin{abstract}
Bayesian optimization for Hyperparameter Optimization (HPO) and Combined Algorithm Selection and Hyperparameter Optimization (CASH) often relies on explicit hyperparameter encodings as surrogate inputs.
In high-dimensional search spaces, however, such representations can make it difficult for standard surrogates to learn useful performance models.
We instead study an alternative meta-learning representation that describes each configuration directly as its historical performance across prior tasks. 
This representation can be paired directly with standard surrogate models.
Our scope is the iterative selection component of AutoML, deciding which configuration to evaluate next from a fixed candidate set, rather than a complete framework with features such as runtime estimation, cost-aware budgeting and ensembling.
We further apply this representation to maximal marginal relevance (MMR), a diversity-based re-ranking criterion from recommender systems, as a posterior-free acquisition rule.
When selecting and optimising across multiple models on the TabArena benchmark, the performance-based representation paired with a linear Bayesian ridge regression model outperforms dedicated meta-learning baselines, achieving the same tuned performance in roughly half the number of trials. 
For single-model HPO, it remains a competitive second. Across both benchmarks this method closely matches or improves on regret at a fraction of the complexity and cost, requiring no pre-training or tuning and scaling independently of search space dimensionality. Code for reproducing our results is available at \url{https://github.com/EWitting/simple-metalearning}.
\end{abstract}

\section{Introduction}\label{sec:intro}

Automated machine learning (AutoML) aims to remove the trial-and-error of selecting an algorithm and tuning its hyperparameters for a given task. The dominant paradigm casts this as hyperparameter optimization (HPO), or, when the algorithm itself is a decision variable, as Combined Algorithm Selection and Hyperparameter Optimization (CASH) \citep{thornton2013auto}. Both are typically solved with Bayesian Optimization (BO), in which a surrogate model predicts the performance of a candidate configuration and an acquisition function selects what to evaluate next \citep{brochu_tutorial_2010}.
When historical results on related tasks are available, meta-learning can transfer this information to improve optimization on a new task \citep{brazdil_metalearning_2022}.
Such transfer has been realised through multi-task and transfer surrogates that share information across related optimization problems \citep{swersky2013multi, wistuba2018scalable, feurer2018practical, wistuba2021few, jomaa2021transfer, wang2024pre}, or more recent meta-trained or in-context models that learn from collections of prior tasks end-to-end \citep{wei2019transferable, ruhkopf2023masif, arangoquick}.

Despite their differences, most of these methods share the same configuration representation: an encoding of the hyperparameter vector, mapped to performance prediction. In CASH, this representation often concatenates the hyperparameter spaces of every candidate algorithm together with an algorithm identifier, leading to a  high-dimensional conditional space in which most input dimensions are inactive for any given configuration \citep{thornton2013auto, feurer2015efficient, lindauer2022smac3}.
Thus, learning useful interactions with such an input space is challenging.

\begin{figure}[t]
    \centering
    \includegraphics[width=0.85\linewidth]{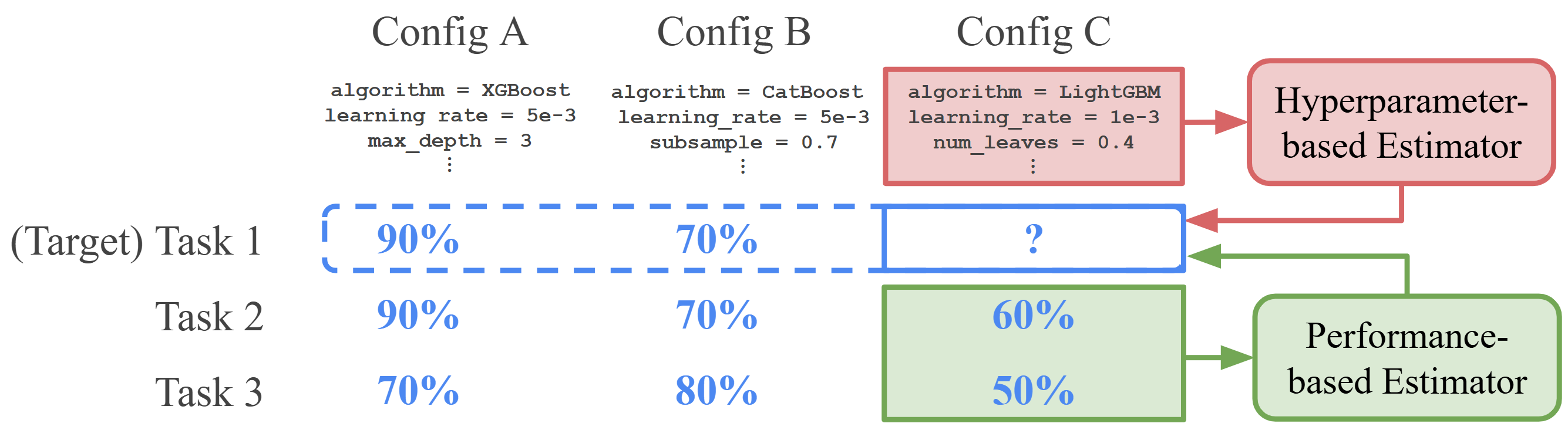}
    \caption{\textbf{Hyperparameter-based vs.\ performance-based configuration representations.} To estimate performance on a target task, standard BO represents a configuration by its hyperparameter vector $\lambda$ (top) and possibly uses results on prior tasks for meta-learning, whereas our approach uses the historical performance vector $P_\lambda$ across prior tasks directly as input representation (bottom). Our approach can be interpreted as learning to describe the target task in terms of prior tasks, for example as weighted sum of prior tasks in the case of a linear estimator.}
    \label{fig:pull}
\end{figure}

An alternative line of work, inspired by collaborative filtering \citep{koren2009matrix}, represents configurations by their historical performance in prior tasks to tackle the AutoML problem \citep{fusi2018probabilistic, yang2019oboe}. This view aligns naturally with recommender systems, where tasks play the role of users, configurations of items, and validation scores of ratings. Prior matrix-based methods typically exploit this structure through compressed or latent representations derived from the task--configuration performance matrix \citep{fusi2018probabilistic,yang2019oboe}.

We instead ask whether the raw historical performance vector itself can serve directly as a configuration representation for standard surrogate models. Figure~\ref{fig:pull} illustrates this contrast between the standard hyperparameter-based view and our performance-based representation. Additionally, we find that historical performance vectors also support a useful measure for representing diversity of configurations. In combination with maximal marginal relevance (MMR), a classical re-ranking criterion from information retrieval and recommender systems \citep{carbonell1998use}, and using historical mean performance as the relevance, this yields MeanMMR, a model-free acquisition rule that can stand in for uncertainty-driven exploration normally seen in Bayesian optimisation.

We benchmark these methods on TabArena \citep{erickson2026tabarena}, a recent benchmark that covers diverse algorithms with randomly sampled hyperparameter configurations each across carefully curated realistic tabular classification and regression tasks, yielding a high-dimensional CASH search space. The performance-based representation paired with Bayesian ridge regression outperforms dedicated meta-learning baselines on CASH and remains a competitive second on HPO, at a fraction of their complexity and cost. Surprisingly, our simple model-free MeanMMR baseline also obtains lower average regret than these baselines on CASH.

\paragraph{Scope.}
We study the selection loop of AutoML, not a complete AutoML framework. Systems such as AutoGluon \citep{erickson_autogluon-tabular_2020} and auto-sklearn \citep{feurer2022auto} add features such as runtime estimation, cost-aware budget allocation and ensembling, which would confound a comparison that isolates the configuration representation, so we do not compare against them. Our representation is a component such frameworks could adopt for their selection step.

To summarize, our contributions are:
\begin{itemize}
    \item We propose to use the raw historical performance vectors directly as configuration representations in BO, offering a simple route to meta-learning that allows standard surrogate models to meta-learn from previous tasks without specialized meta-learning procedures.
    \item We introduce MMR as a diversity-based, posterior-free acquisition function, and show that combined with historical mean performance it forms MeanMMR, a strong model-free meta-learning baseline.
    \item We benchmark popular surrogate models crossed with both representations on the TabArena benchmark, isolating the effect of the representation, and find that the performance representation makes a simple Bayesian ridge surrogate competitive with or better than dedicated meta-learning baselines at a fraction of their cost.
\end{itemize}

\section{Related Work}
Table~\ref{tab:positioning} provides an overview of related work and the positioning of our work.

\paragraph{Hyperparameter Optimization and CASH.}
HPO aims to find the hyperparameter configuration of a learning algorithm that maximizes a chosen performance metric on a target dataset \citep{hutter2019automated}.
\citet{thornton2013auto} extend HPO to the case where the algorithm itself is also a decision variable, Combined Algorithm Selection and Hyperparameter Optimization (CASH).
As the number of hyperparameters grows, exhaustive enumeration becomes infeasible, and random search becomes a simple and widely used baseline \citep{bergstra2012random,hutter2019automated}.

To handle large search spaces, practical HPO and CASH methods treat configuration search as a sequential allocation problem, deciding which candidate to evaluate next under a limited evaluation budget.
\citet{hutter2011sequential} introduce Sequential Model-Based Optimization (SMBO), which fits a surrogate model to observed configuration performances and uses an acquisition function to select the next configuration.
Bayesian Optimization (BO) \citep{snoek_practical_2012} is a common probabilistic instance of this framework, using the surrogate's predictive distribution to trade off exploiting promising configurations and exploring uncertain regions \citep{brochu_tutorial_2010}.
Modern AutoML systems typically operate on explicit hyperparameter-space encodings of candidate configurations \citep{lindauer2022smac3,falkner2018bohb,thornton2013auto,kotthoff2017auto,feurer2015efficient,feurer2022auto}.

The CASH search space is structurally more challenging than ordinary HPO, concatenating hyperparameters across every candidate algorithm into a high-dimensional, heavily conditional space \citep{feurer2015efficient, lindauer2022smac3}.
Prior work also finds that only a small subset of hyperparameters tends to dominate performance variation \citep{hutter2014efficient, van2018hyperparameter, probst2019tunability}, so a surrogate using an explicit hyperparameter encoding must infer the relevant dimensions and interactions from few target-task evaluations.

\paragraph{Algorithm Selection.}
A related problem setting is algorithm selection (AS) from a finite portfolio of algorithms, without explicitly considering any hyperparameter values \citep{brazdil_metalearning_2022}. The simplest strategy ranks candidates by their average performance across prior tasks. Active Testing \citep{leite_selecting_2012} makes this adaptive: it selects which candidate to evaluate next on the target task by estimating, from cross-task performance similarity, which evaluation is most informative for identifying the best candidate. These methods rely entirely on historical performance rather than a hyperparameter-space surrogate, similar to our setting, but use fixed or heuristic tournament-style ordering without Bayesian optimisation's uncertainty-driven exploration. We instead reframe HPO and CASH as the algorithm selection problem setting, while retaining Bayesian optimisation techniques.

\paragraph{Meta-Learning and Transfer for Bayesian Optimization.}
When historical results on related tasks are available, Bayesian optimization can transfer information to a new task \citep{brazdil_metalearning_2022}. \citet{swersky2013multi} introduce multi-task BO, which couples related problems through a joint task kernel. Ranking-weighted ensembles such as RGPE \citep{feurer2018practical} and two-stage transfer surrogates such as TST-R \citep{wistuba2018scalable} combine per-task Gaussian process models using cross-task ranking agreement. ABLR \citep{perrone_scalable_2018} meta-trains a shared hyperparameter feature embedding network with a per-task Bayesian linear head. Further related work spans deep-kernel transfer, pre-trained Gaussian processes, and in-context meta-models \citep{wistuba2021few,jomaa2021transfer,wang2024pre,wei2019transferable,arangoquick,ruhkopf2023masif}. These methods share two traits: a hyperparameter-encoded configuration input, and a need for specialized meta-learning machinery. We instead move the meta-learning into the input representation, so that a standard surrogate suffices.

\paragraph{Collaborative Filtering and Matrix-Based AutoML.}
A parallel line of work, inspired by collaborative filtering, observes that a task-configuration performance matrix mimics the user-item ratings matrix used in recommender systems. Tasks act as users, configurations as items, and validation scores as ratings \citep{koren2009matrix,yang2019oboe,fusi2018probabilistic}. Existing matrix-based methods typically compress this matrix into latent representations. \citet{yang2019oboe} introduce OBOE, which constructs compact configuration representations using low-rank matrix factorization and PCA-based embeddings. \citet{fusi2018probabilistic} introduce PMF, which learns latent configuration representations by gradient-based optimization and naturally handles sparse performance matrices. Two-Tower models add deep learning to factorization, generating separate task and configuration embeddings via multilayer perceptrons before scoring their compatibility \citep{huang2013learning,covington2016deep,yi2019sampling,raza2026comprehensive}. In contrast to all of these, we use the observed historical performance vector directly as a surrogate input, without an additional compression or factorization step. Our linear model is related to OBOE, and we compare Two-Tower variants embedding either hyperparameters or performance directly (see Appendix~\ref{app:two-tower-tuning}).

\subsection{Diversity-Based Acquisition and MMR}

Beyond the surrogate model, our work also revisits the acquisition function used to select the next configuration.
A common BO acquisition function is Expected Improvement (EI) \citep{jones1998efficient}, which uses the surrogate's predictive posterior to balance promising predicted performance against uncertainty (see Appendix~\ref{app:acquisition_details} for a short primer).
When the surrogate does not provide a calibrated predictive posterior, standard uncertainty-driven rules such as EI do not directly apply. Prior matrix-based AutoML methods face the same issue: OBOE's bilinear low-rank surrogate \citep{yang2019oboe} is not posterior-providing, so it replaces EI with a custom procedure, motivating more general posterior-free alternatives.

Maximal marginal relevance (MMR) was originally introduced for document retrieval and summarization and has since been used as a diversification strategy in recommender and retrieval-oriented settings \citep{carbonell1998use,vargas2011rank,di2017adaptive,chen2021multi,do2025considering,korikov2026rebol}. Related diversity-aware ideas have also appeared in ensemble-oriented CASH and diversified batch Bayesian optimization \citep{shen2022divbo,nava2022diversified}. However, to the best of our knowledge, we are the first to formulate MMR as acquisition rule for diversity-aware exploration in standard HPO and CASH, providing a simple posterior-free alternative to uncertainty-driven acquisition.

\begin{table}[t]
\centering
\caption{Positioning of our approach against representative methods from four lines of related work. We combine the performance-matrix representation of algorithm selection with the active surrogate-and-EI loop of Bayesian optimization, using the raw performance vector and any standard surrogate, without bespoke meta-training or latent compression. Abbreviations: BO, Bayesian optimization; AS, algorithm selection; CF, collaborative filtering.}
\label{tab:positioning}
\begingroup
\small
\setlength{\tabcolsep}{3pt}
\begin{tabular}{>{\raggedright\arraybackslash}p{4.45cm}>{\centering\arraybackslash}p{2.6cm}>{\centering\arraybackslash}p{2.15cm}>{\centering\arraybackslash}p{2.25cm}>{\centering\arraybackslash}p{1.35cm}}
\toprule
\textbf{Method} & \textbf{Line of work} & \textbf{Config.\ representation} & \textbf{Surrogate} & \textbf{Selection} \\
\midrule
SMAC3 \citep{lindauer2022smac3}                & BO & Hyperparameter        & Random forest          & EI \\
GP-BO \citep{snoek_practical_2012}             & BO & Hyperparameter        & Gaussian process       & EI \\
\addlinespace
Average ranking \citep{brazdil_metalearning_2022} & AS & Performance   & None (ranking)         & Static \\
Active Testing \citep{leite_selecting_2012}    & AS   & Performance    & None (matrix)          & Active \\
\addlinespace
Multi-task GP \citep{swersky2013multi}         & Meta-learning BO           & Hyperparameter        & Multi-task GP          & EI \\
RGPE \citep{feurer2018practical}               & Meta-learning BO           & Hyperparameter        & GP ensemble            & EI \\
TST-R \citep{wistuba2018scalable}              & Meta-learning BO           & Hyperparameter        & GP ensemble            & EI \\
ABLR \citep{perrone_scalable_2018}             & Meta-learning BO           & Hyperparameter        & NN $+$ Bayesian linear & EI \\
\addlinespace
OBOE \citep{yang2019oboe}                       & CF / matrix           & Performance (PCA)     & Linear                 & Custom \\
PMF \citep{fusi2018probabilistic}               & CF / matrix           & Performance (latent)  & Gaussian process       & EI \\
Two-Tower \citep{huang2013learning}             & CF / matrix           & Hyper.\ / perf.\ (latent) & Two-tower network  & Greedy / MMR \\
\midrule
\textbf{Ours}                                   & \textbf{AS $+$ BO} & \textbf{Performance} & Any regression model & EI / MMR \\
\bottomrule
\end{tabular}%
\endgroup
\end{table}

\section{Performance-Based Representations for Simple Meta-Learning}
\label{sec:method}

\subsection{Problem Setup}
\label{sec:setup}

We study HPO and CASH as sequential configuration selection over a finite candidate set $\Lambda$.
Each $\lambda \in \Lambda$ denotes a fully specified candidate that can be evaluated on a learning task $t$, yielding a performance score $\mathcal{P}(t, \lambda)$ where larger values indicate better performance.
In the HPO setting, $\Lambda$ contains configurations of a single algorithm; in the CASH setting, $\Lambda$ spans a joint conditional configuration space over all candidate algorithms and their hyperparameters, represented through a one-hot algorithm indicator together with the full hyperparameter encoding.
For a target task $t$, an optimization method iteratively selects an unevaluated $\lambda \in \Lambda$, observes $\mathcal{P}(t,\lambda)$, and updates a surrogate estimator $\hat{\mathcal{P}}$ for unobserved configurations.
We restrict all methods to the same $\Lambda$ and instantiate each surrogate model family with two alternative inputs $x_\lambda$: the standard hyperparameter-based representation and a performance-based representation built from the historical performance across prior tasks, allowing simple meta-learning.

\subsection{Performance-Based Representation for Meta-Learning}

We propose a performance-based representation that describes a configuration by how it performed on previous tasks, instead of describing a configuration in terms of hyperparameter values.
Let $T_{\mathrm{train}}$ be the set of prior tasks available for meta-learning.
We collect the corresponding scores in a historical performance matrix $P \in \mathbb{R}^{|T_{\mathrm{train}}| \times |\Lambda|}$, where each entry is $P_{t, \lambda} = \mathcal{P}(t, \lambda)$ for $t \in T_{\mathrm{train}}$ and $\lambda \in \Lambda$.
Each column of $P$ corresponds to one configuration and each row corresponds to one prior task.
The column $P_\lambda := P_{:, \lambda}$ is therefore the historical performance vector of configuration $\lambda$, and we use this column as the estimator input, $x_\lambda^{\mathrm{perf}} = P_\lambda$.
Configurations are thus compared by their empirical behaviour across previous tasks rather than by their hyperparameter values.
This moves meta-learning into the input representation itself, allowing any standard surrogate that accepts vector inputs to exploit previous tasks by replacing $x_\lambda^{\mathrm{hyper}}$ with $x_\lambda^{\mathrm{perf}}$, without changing the surrogate architecture, kernel, or training objective.

\subsection{Surrogate Models for Performance Based Representations}
\label{sec:surrogates}
We focus on the surrogate central to our contribution, Bayesian ridge regression, which to the best of our knowledge we are the first to study for HPO and CASH with the proposed performance-based representation. We additionally study a Two-Tower collaborative-filtering network as a deep meta-learning comparison; as it is not a central contribution, its description and results are deferred to Appendix~\ref{app:two-tower-tuning}.

\paragraph{Linear (Bayesian ridge regression).}
Since the performance-based representation already encodes how each configuration behaves across prior tasks, we use Bayesian ridge regression \citep{mackay1992bayesian,tipping2001sparse} as a low-capacity surrogate that remains stable with few target-task observations and provides the predictive posterior needed for EI \citep{jones1998efficient}. Specifically, we use a linear-Gaussian model with a zero-mean isotropic Gaussian prior on the weights and Gamma hyperpriors on the noise and weight precisions. Training uses iterative empirical Bayes/evidence-maximization updates of these precisions rather than variational inference; conditional on the precisions, the weight posterior is Gaussian in closed form. When the input is $x_\lambda^{\mathrm{perf}}$, the coefficients can be interpreted as weights over prior tasks in $T_{\mathrm{train}}$, so target-task adaptation learns which previous tasks are informative for the current one. Implementation details are given in Appendix~\ref{app:bayesian_ridge}, and its relation to OBOE is discussed in Appendices~\ref{app:representation_encoding} and~\ref{app:linear_surrogate_choice}.

\subsection{The maximal marginal relevance (MMR) Acquisition and MeanMMR Baseline}
\label{sec:mmr_acquisition}

To support surrogates without a calibrated predictive posterior, we adapt MMR \citep{carbonell1998use} as a diversity-based acquisition function that selects configurations with high predicted relevance while penalizing similarity to configurations already evaluated on the target task. Let $\Lambda_{\mathrm{obs}}$ be the set of configurations already evaluated on the target task. MMR selects the next configuration by maximizing:
\begin{equation}\label{eq:mmr}
    \mathrm{MMR}(\lambda) = \alpha \cdot \hat{\mathcal{P}}(t, \lambda) - (1 - \alpha) \cdot \max_{\lambda' \in \Lambda_{\mathrm{obs}}}\, \cos(z_\lambda,\, z_{\lambda'})
\end{equation}
where $\hat{\mathcal{P}}(t,\lambda)$ is the relevance score, $z_\lambda$ is the representation used to compute similarity, and $\alpha$ controls the relevance-diversity tradeoff (larger $\alpha$ places more emphasis on relevance, while smaller $\alpha$ encourages greater novelty relative to previously evaluated configurations).

Because MMR only requires a relevance score and a similarity, we can drop the surrogate altogether. MeanMMR replaces the relevance score $\hat{\mathcal{P}}(t, \lambda)$ in Equation~\ref{eq:mmr} with the historical mean performance of $\lambda$ across $T_{\mathrm{train}}$, and computes similarity directly on the performance vectors $P_\lambda$ (full expression in Appendix~\ref{app:acquisition_details}).
It turned out to be a surprisingly strong baseline.

\begin{figure}[t]
  \centering
  \includegraphics[width=\textwidth]{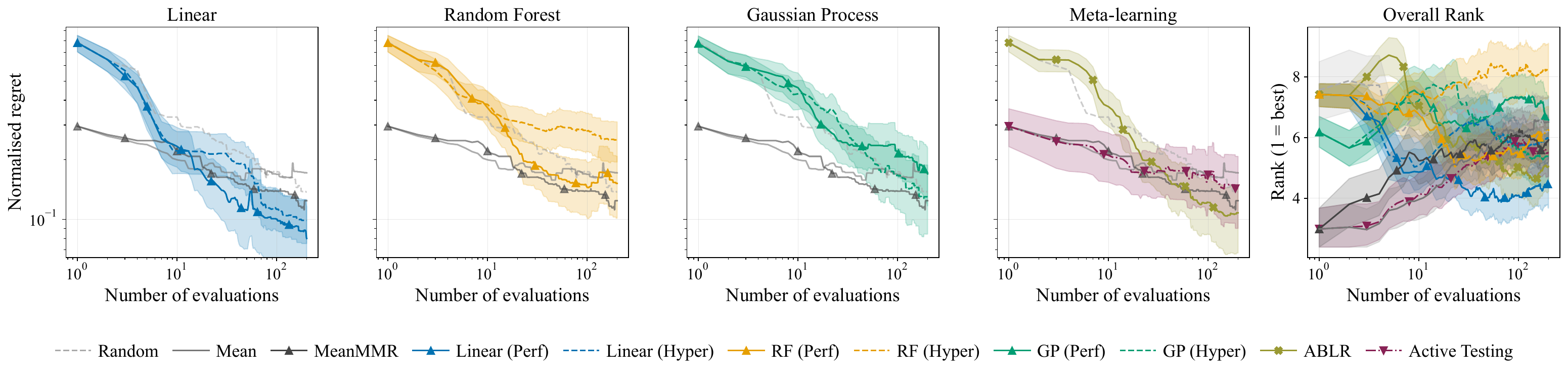}
  \caption{Test normalized regret per model family and overall rank vs.\ number of evaluations
    on the CASH benchmark (TabArena). Methods introduced in this work (the Perf variants and MeanMMR) are marked with an upward-pointing triangle ($\blacktriangle$). Shaded bands are 95\% bootstrap CIs across tasks.}
  \label{fig:cash_results}
\end{figure}

\section{Experimental Setup} \label{sec:experiments}

\paragraph{Baselines.}
We compare results against dedicated meta-learning methods: the hyperparameter-based MTGP \citep{swersky2013multi}, RGPE \citep{feurer2018practical} and ABLR \citep{perrone_scalable_2018}, as well as Active Testing \citep{leite_selecting_2012} designed for algorithm selection. These dedicated baselines are included where our compute budget allowed, as the GP-based methods (MTGP, RGPE) do not stay computationally feasible on the high-dimensional CASH benchmark, and MTGP required subsampling modification to apply to HPO (see Appendix~\ref{app:gp-cost} for computational cost analysis).

In addition, we include three more reference baselines. Random search \citep{bergstra2012random} uses no prior information and suggests configurations in random order. Mean ranking orders configurations by their average performance in $P$, one of the simplest forms of meta-learning and a well-studied baseline for both HPO and CASH \citep{brazdil_metalearning_2022}. MeanMMR is the model-free acquisition baseline defined in Section~\ref{sec:mmr_acquisition}.

\paragraph{Our models.}
We compare directly the hyperparameter-based and performance-based representations introduced in Section \ref{sec:method} on three surrogate models: Bayesian ridge regression, the ubiquitous Gaussian process regressor \citep{balandat2020botorch}, with automatic relevance determination (ARD) enabled to assign an independent lengthscale per input dimension (implementation details in Appendix~\ref{app:gp-details}), and a Random Forest inspired by SMAC \citep{lindauer2022smac3}.
Throughout the experiments, we use the suffixes Hyper and Perf to denote variants using hyperparameter-based and performance-based representations, respectively.
Within each surrogate family, the two variants differ only in whether the surrogate receives $x_\lambda^{\mathrm{hyper}}$ or $x_\lambda^{\mathrm{perf}}$. We use explicit hyperparameter encodings for the hyperparameter-based variants and the full historical performance vector $P_\lambda$ for the performance-based variants; implementation details are given in Appendix~\ref{app:representation_encoding}.

\paragraph{Dataset.}

All experiments use the TabArena benchmark \citep{erickson2026tabarena} (see Appendix~\ref{app:dataset}), which suits our study by combining curated tabular classification and regression tasks with leaderboard results for many modern algorithms and numerous sampled configurations per algorithm, allowing us to construct both CASH and HPO benchmarks from a shared task-configuration performance matrix.
Specifically, we construct two finite search spaces: a CASH benchmark over a multi-algorithm configuration set (see Appendix~\ref{app:base_models}) and an HPO benchmark over an XGBoost-only subset (see Appendix~\ref{app:xgboost_search_space}).
The candidates in each benchmark define $\Lambda$ and, on prior tasks, the performance matrix $P$ used by the performance-based methods.
Detailed benchmark construction is described in Appendix~\ref{app:benchmark_construction}.

\paragraph{Metrics and Validation Scheme.}\label{sec:validation}
We use five-fold cross-validation at the task level. Meta-training tasks define the performance matrix and any prior-task training, while held-out tasks are optimized independently as target HPO or CASH problems. Search uses validation-fold feedback only, and final regret is computed on separate test folds for the validation-selected configuration to avoid skewed results due to adaptive overfitting \citep{schroder_overfitting_2025}. Further details are given in Appendix~\ref{app:evaluation_protocol}.

\section{Results}

On the high-dimensional CASH benchmark, the performance-based representation paired with a simple Bayesian ridge surrogate (Linear-Perf) attains the best average regret and rank overall, both in the regret curves (Figure~\ref{fig:cash_results}) and at the full budget of 200
evaluations (Table~\ref{tab:main_results}). Concretely, ABLR needs 149 evaluations to reach the regret Linear-Perf reaches in 73, close to a 2x reduction in trials needed. Linear-Perf is also the fastest of the surrogate models to fit, averaging under three seconds per task against roughly two hours for GP-Hyper on the same benchmark. On the lower-dimensional HPO benchmark, the ABLR method that meta-learns a hyperparameter embedding network performs best, with Linear-Perf following at a close second.

\begin{table}[t]
\centering
\caption{Average normalized regret on the test split ($\times 100$) and average rank across tasks, over 200 evaluations amongst the discrete set of configurations for the CASH TabArena benchmark and the HPO TabArena XGBoost-only benchmark.
\textbf{Bold}: best per metric per benchmark.
\underline{Underline}: best per model family per metric.
$^*$: significantly better than others within the family (Wilcoxon signed-rank, $p < 0.05$). Time/task: single-threaded per-task time spent fitting the surrogate and evaluating the acquisition function over the 200 evaluations. $^{\ddagger}$: excludes the required one-time pretraining phase shared across tasks (ABLR 3.4m on CASH, 27s on HPO; RGPE 2.0m on HPO).}
\label{tab:main_results}
\begingroup
\small
\setlength{\tabcolsep}{3pt}
\begin{tabular}{l l ccc ccc}
\toprule
& & \multicolumn{3}{c}{\textbf{CASH }}
  & \multicolumn{3}{c}{\textbf{HPO }} \\
\cmidrule(lr){3-5} \cmidrule(lr){6-8}
\textbf{Model} & \textbf{Input}
  & Regret $\downarrow$ & Rank $\downarrow$ & Time/task
  & Regret $\downarrow$ & Rank $\downarrow$ & Time/task \\
\midrule
\multirow{2}{*}{Linear}
  & Perf.    & \underline{\textbf{12.19}}$^{*}$ & \underline{\textbf{4.34}}$^{*}$ & 2.7s  & \underline{9.84}$^{*}$ & \underline{6.14}$^{*}$ & 1.1s \\
  & Hyper.   & 14.59 & 5.95 & 24s  & 12.79 & 7.04 & 0.7s \\
\addlinespace
\multirow{2}{*}{Random Forest}
  & Perf.    & \underline{18.53}$^{*}$ & \underline{5.83}$^{*}$ & 2.0m & \underline{11.29} & \underline{6.51} & 1.8m \\
  & Hyper.   & 28.59 & 8.06 & 2.7m & 13.46 & 6.59 & 53s \\
\addlinespace
\multirow{2}{*}{Gaussian Process}
  & Perf.    & 23.73 & 6.99 & 7.3m & \underline{13.40} & 6.87 & 1.5m \\
  & Hyper.   & \underline{20.25} & \underline{5.93} & 1.8h & 13.61 & \underline{6.68} & 7.9m \\
\midrule
ABLR           & Hyper. & \underline{15.47} & \underline{5.22} & 1.1s$^{\ddagger}$ & \underline{\textbf{9.02}} & \underline{\textbf{5.97}} & 0.6s$^{\ddagger}$ \\
Active-Testing & Perf.  & 16.54 & 5.37 & 0.9s & 16.63 & 7.56 & 0.7s \\
MTGP           & Hyper. & --- & --- & --- & 13.52 & 6.98 & 10s \\
RGPE           & Hyper. & --- & --- & --- & 10.57 & 6.18 & 9.3m$^{\ddagger}$ \\
\midrule
Random   & ---      & 19.45 & 6.61 & 0.5s & 16.86 & 7.63 & 0.3s \\
Mean     & Perf.    & 17.56 & 5.93 & 0.7s & 25.03 & 9.37 & 0.3s \\
MeanMMR  & Perf.    & \underline{14.60} & \underline{5.77} & 6.6s & \underline{14.92} & \underline{7.47} & 0.5s \\
\bottomrule
\end{tabular}%
\endgroup
\end{table}

Comparing the performance representation against the usual hyperparameter representation shows statistically significant improvements for the linear and random forest surrogates in the CASH benchmark, but no statistically significant or consistent improvement for the Gaussian Process: the benefit of historical performance as representation is not observed as strongly across all surrogate types.

Rankings of the static baselines that do not fit a surrogate model differ between benchmarks. On the XGBoost HPO setting, all model-based methods consistently achieve better regret than random search after the first few iterations, with mean ranking the worst strategy overall ($25.03$ versus
$16.86$ for random search). In the CASH setting, the hyperparameter-based random forest does not clearly improve upon random search, and the mean ranking gives a very strong initial selection of models until being surpassed after ten evaluations, and still beats random search (regret $17.56$ versus $19.45$). Active testing benefits from the strong initialization on CASH, and is surpassed by Linear-Perf after roughly twenty evaluations. The MeanMMR baseline improves upon mean ranking, such that it beats random search on HPO, and is competitive with non-metalearning methods for roughly the first fifty evaluations, although surpassed earlier by strong meta learners. Surprisingly, MeanMMR outperforms the dedicated meta-learning methods on CASH in terms of average regret.

\begin{figure}[tb]
  \centering
  \includegraphics[width=0.85\linewidth]{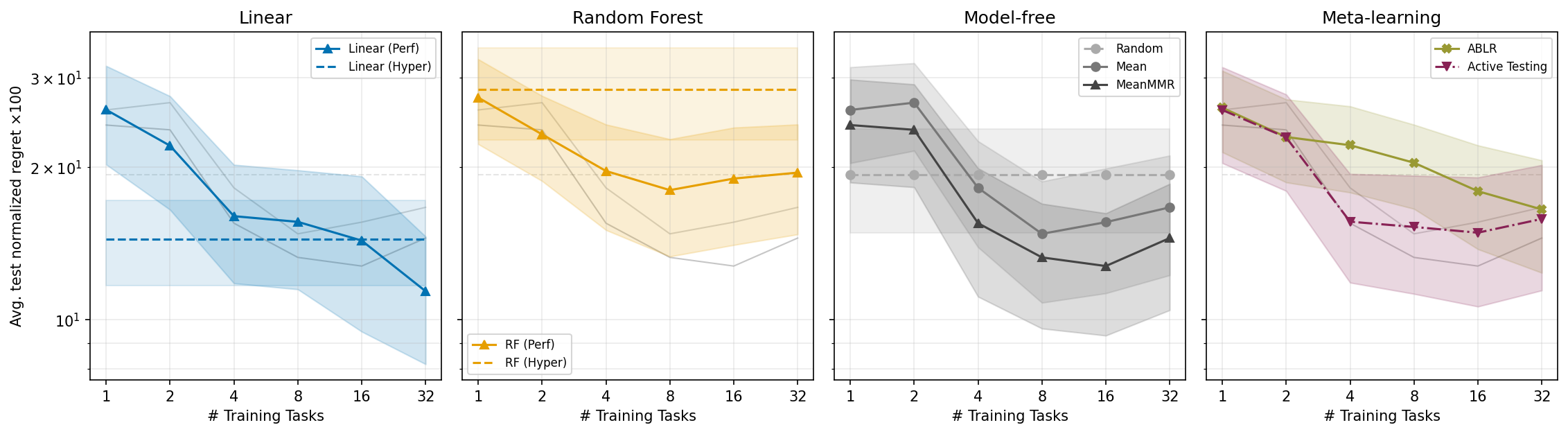}
  \caption{Average normalized regret over 200 configuration evaluations on the CASH benchmark vs. number of training tasks. Methods introduced in this work (the Perf variants and MeanMMR) are marked with an upward-pointing triangle ($\blacktriangle$). Dashed horizontal lines show methods that do not make use of results on previous tasks.
    Shaded bands are 95\% bootstrap CIs across tasks.}
  \label{fig:scalability_cash}
\end{figure}

We measure how the methods depend on the number of tasks for which historical performance is known (Figure~\ref{fig:scalability_cash}); as expected, average regret generally improves as the number of prior tasks increases. For CASH, Linear-Hyper is the strongest non-metalearning method whose average regret is not surpassed until at least 16 training tasks are used by Linear-Perf. In the HPO setting, ABLR is immediately competitive with Linear-Hyper and offers the best regret once results for at least one prior training task are available. Methods using mean ranking, including active testing, appear to benefit initially but stop improving after more than 16 tasks are added.

\section{Discussion \& Limitations}\label{sec:discussion}

The performance-based representation implements meta-learning well specifically in combination with a linear surrogate model, indicating that a configuration's performance on the target task fits well to a linear combination of its performances on prior tasks: Linear-Perf effectively learns to represent the target task as a weighted average of prior tasks. That aligns with ideas behind related work such as the weighted ensembling over task models of RGPE. We performed an ablation of RGPE that replaces the base-GP predictions with the true performance columns of the prior tasks (RGPE-Oracle), which is still strong on HPO but not as competitive with Linear-Perf on CASH (see Appendix~\ref{app:gp-cost}). Freely determining weights per prior task via regression may be more effective and expressive than RGPE's non-negative heuristic ranking agreement; a fixed-precision ablation confirms this is not simply due to adaptive precision estimation (see Appendix~\ref{app:bayesian_ridge}). Consistent with this, the weights learned by Bayesian ridge regression tend to be both positive and negative, with roughly half negative on CASH (see Appendix~\ref{app:weight_signs}), likely necessary for a good fit on the high-dimensional problem.

Rankings of models differ between the CASH and HPO settings. In the lower-dimensional HPO case, hyperparameter methods all perform better than random search by a larger margin, and the hyperparameter embedding-based ABLR method is best overall. This can be explained by the manageable dimensionality, with only 11 hyperparameters whose effects to learn, versus 475 values after encoding for the high-dimensional CASH problem. This reflects the curse of dimensionality: more data points are needed to obtain the same predictive performance, and since only a subset of hyperparameters is active per selected algorithm, relevant data points per hyperparameter are further reduced. Architectures that exploit this structure efficiently, e.g., separate encoders per algorithm or surrogates designed for conditional, tree-structured search spaces \citep{jenatton2017bayesian}, might have performed better; not pursuing that direction is a limitation of our baselines, but consistent with our motivation that large-scale CASH invites unwanted engineering complexity. One might argue this large-scale task is an unfair comparison, but it also reflects a realistic problem: the choices data scientists face with the full breadth of state-of-the-art tabular models on benchmark leaderboards such as TabArena \citep{erickson2026tabarena}.

The model-free baseline results also support that meta-learning helps more on CASH than on HPO. Within one algorithm's hyperparameter space the choice of configuration might depend more on the specific tasks, while across multiple algorithm types some might be clearly stronger or weaker. That might explain why historical performance gives larger advantages on CASH.

The strong initialisation of the mean best configuration on CASH, as used by active testing, could easily be applied to other methods as well, further strengthening for example Linear-Perf. This was not done for the base model set to keep the experiment controlled by only varying the input representation, and the other meta-learning baselines did not explicitly include this initialisation step in their algorithm.

MeanMMR benefits from the strong mean signal, while not suffering from the low performance that mean ranking has on HPO. The exploration term based on historical performance diversity bridges part of the gap between static ranking methods and exploration-driven Bayesian optimisation. The low performance of mean ranking on HPO might therefore also be caused by it suggesting low-diversity configurations from a local optimum, suggesting portfolio methods might benefit from similar diversity measures.
MMR also extends to surrogate models without a predictive posterior, such as the Two-Tower recommender model; we explored this variant but omit it from the main comparison as its results were competitive but not state-of-the-art (see Appendix~\ref{app:two-tower-tuning}).

The biggest limitation of the performance-based methods is requiring a dense matrix with historical performance of every candidate configuration on every prior task. The cost of collecting the matrix scales with both dimensions, number of configurations and number of tasks. Given this initial cost, the method is most applicable to popular search spaces where a centralized matrix can be maintained, shared and reused many times for different applications. For niche domains and architectures where a search space is only used a handful of times, benefits of faster search likely do not outweigh the initial cost of collecting the matrix. Furthermore, unlike factorization methods such as PMF \citep{fusi2018probabilistic}, our method would require imputation to support sparse matrices.

Multi-task benchmarks give a more optimistic perspective on this data requirement. In our scalability experiment on CASH, roughly 16 prior tasks were needed before performance representation models outperformed alternatives, while the full TabArena leaderboard \citep{erickson2026tabarena} already contains results for 51 tasks. Matrices built from benchmark leaderboards are also likely to stay up to date, since authors of new methods contribute the required data whenever they report results on the benchmark.

Scaling of possible performance with respect to the number of configurations is clear: A performance-based method can at best suggest the optimal configuration included in the matrix, when the truly optimal configuration may in reality lie somewhere else in the continuous hyperparameter search space. Future work should examine the continuous case.

Because it requires no hyperparameter encoding at all, our representation also avoids having to define a single unified search space across heterogeneous algorithms, which is especially useful when aggregating results across many different model types, for example from platforms such as OpenML \citep{bischl2025openml}.

\section{Conclusion}
In summary, using historical performance as a configuration representation is an effective and simple approach to meta-learning. 
Paired with a linear Bayesian ridge regression model, it outperforms dedicated meta-learning baselines on the high-dimensional CASH benchmark, reaching the same tuned performance in roughly half the number of trials, and remains a competitive second on single-model HPO.
Simplicity is key: the well-performing Bayesian ridge regression only models linear relationships between tasks and requires no pre-training, hyperparameter preprocessing, or tuning. Using performance representations avoids complexity in combining search space encodings across models with different hyperparameters. For a completely model-free alternative, ranking by the mean and diversity of the performance vectors (MeanMMR) is stronger than pure average ranking and competitive with limited target-task observations.

\section*{AI use statement}

We have not used generative AI tools for any of the following tasks: generating synthetic data sets, developing theoretical models or conceptual frameworks, formulating mathematical claims, providing critical ingredients for proving mathematical claims, assisting in the writing of proofs, proposing or refining hypotheses, designing or providing feedback on research methodology or experiments, implementing methods, assisting with translation, cleaning and reformatting data sets, supporting qualitative and thematic data analysis, and interpreting results. We did use generative AI tools for sourcing and summarising existing literature, for editing software code, limited to refactoring, visualisation scripts, and scripts that aggregate experiment outputs into tables and run the reported significance tests, for creating and modifying figures, for suggesting a structure for this paper, for \LaTeX{} formatting support, and for proofreading to improve readability. We have reviewed all AI-assisted work: all reported results were produced and verified by the authors' own code, AI-suggested references were checked against the primary sources, and all AI-assisted text was read and edited by the authors. We take responsibility for the final content of this work, including text, claims or artifacts produced with the aid of generative AI.

\section*{Reproducibility statement}

All reported experiments are simulated by lookups into the public TabArena leaderboard results \citep{erickson2026tabarena}, so no base model has to be retrained to reproduce them. Appendix~\ref{app:benchmark_construction} describes how the CASH and HPO search spaces are constructed from that data, Appendix~\ref{app:dataset} lists the exact base model set and the XGBoost search space, and Appendices~\ref{app:evaluation_protocol} and~\ref{app:regret_normalization} define the task-level cross-validation protocol, the validation-test split, and the regret normalization. Implementation details for every surrogate and baseline are given in Appendices~\ref{app:representation_encoding}--\ref{app:two-tower-tuning}, with the final hyperparameters and random seeds collected in Table~\ref{tab:final_hyperparams}, and the hardware, software versions, and total compute are reported in Appendix~\ref{app:compute}. The source code, including the configuration files that define each reported run, is available at \url{https://github.com/EWitting/simple-metalearning}.

\bibliography{mylibrary}
\bibliographystyle{iclr2027_conference}

\clearpage
\appendix

\section*{Appendix}

This appendix provides the implementation and evaluation-protocol details, additional figures, and extended discussion referenced from the main paper.

\section{HPO Benchmark Figures} \label{app:hpo benchmark}
\begin{figure}[htb]
  \centering
  \includegraphics[width=\linewidth]{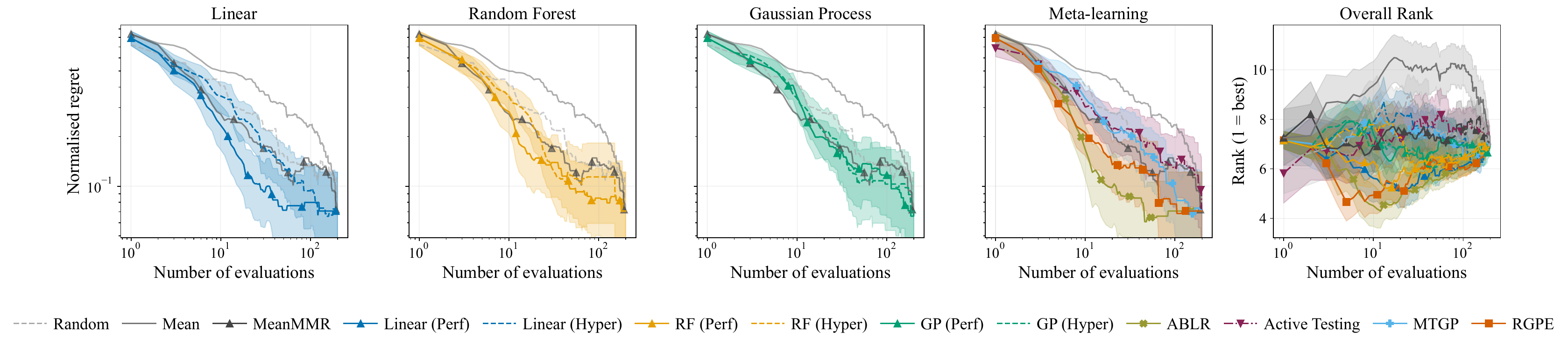}
  \caption{Test normalized regret per model family and overall rank vs.\ number of evaluations
    on the HPO benchmark (TabArena XGBoost-only). Methods introduced in this work (the Perf variants and MeanMMR) are marked with an upward-pointing triangle ($\blacktriangle$). Shaded bands are 95\% bootstrap CIs across tasks.}
  \label{fig:hpo_results}
\end{figure}

\begin{figure}[htb]
  \centering
  \includegraphics[width=\linewidth]{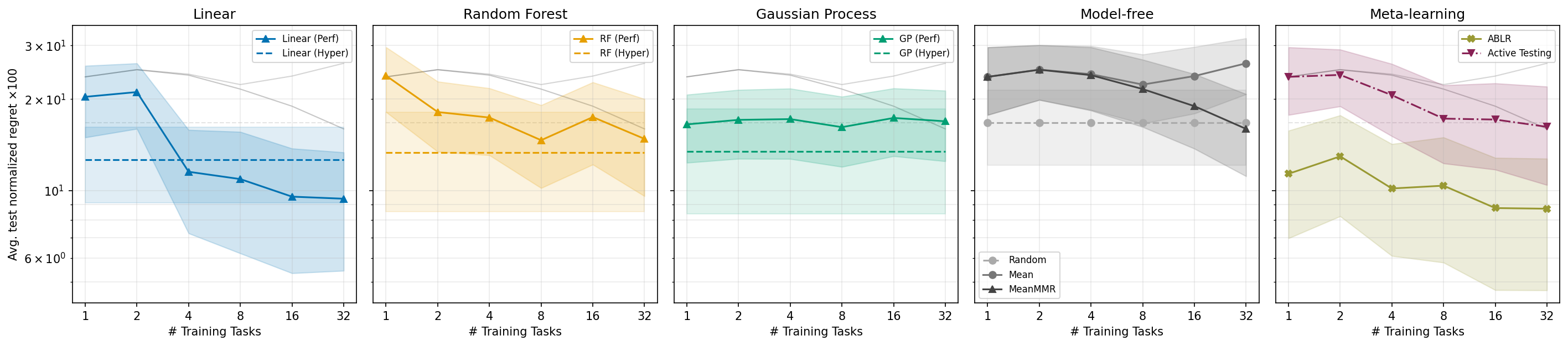}
  \caption{Average normalized regret over 200 model evaluations on the HPO benchmark vs.\
    number of previous tasks observed, grouped by model family. Methods introduced in this work (the Perf variants and MeanMMR) are marked with an upward-pointing triangle ($\blacktriangle$).
    Dashed horizontal lines show methods that do not make use of results on previous tasks.
    Shaded bands are 95\% bootstrap CIs across tasks.}
  \label{fig:scalability_hpo}
\end{figure}

Plots showing the test normalized regret against number of evaluations or number of observed prior tasks on the HPO benchmark are shown in Figure~\ref{fig:hpo_results} and Figure~\ref{fig:scalability_hpo}.

\section{Experimental Setup Details}\label{app:experiment_setup_details}
This appendix provides the implementation details behind the experimental protocol, including benchmark construction, encoding choices, acquisition rules, and model settings.

\subsection{Benchmark Construction}\label{app:benchmark_construction}
The public TabArena leaderboard dataset contains cross-validated results for evaluating each listed algorithm on each task.
For most algorithms this includes not only the default hyperparameter setting, but also 200 randomly selected hyperparameter configurations.
From these data we construct the performance matrix $P$ used in our meta-learning models and representations.
Using all base algorithms and tasks, the matrix contains 3623 configurations across 23 algorithms and 51 tasks (see Table~\ref{tab:base_models}). The algorithm set does not include large ensembles like AutoGluon, as they could have a disproportionate impact through their high performance, while our methods and evaluation framework do not penalize them for disproportionately high training cost.
For the HPO benchmark, we use the 201 XGBoost configurations, consisting of one default configuration and 200 random configurations (see Table~\ref{tab:xgboost_search_space}).
The choice of XGBoost is arbitrary; any competitive method with a sufficiently large list of configurations could have been used.

These 3623 configurations for CASH and 201 configurations for HPO fully define the search space $\Lambda$ in each benchmark.
Such a finite set is required for defining the matrix used by performance-based methods.
The hyperparameter-based methods do not have that restriction in principle and could suggest configurations from the underlying continuous search space.
However, using the same finite search space for all variants isolates the effect of changing the representation.
Additionally, staying within this set makes large-scale experimentation possible by simulating configuration evaluations with lookups in the precomputed benchmark results.

\subsection{Evaluation Protocol}\label{app:evaluation_protocol}
We employ five-fold cross-validation at the task level, each fold resulting in a \textit{meta-training set} of approximately 40 training tasks and 10 test tasks.
The meta-training tasks are used for the performance matrix and representations, and for training the Two-Tower model, though not all methods make use of them.
Within the meta-test set, HPO or CASH is run on each test task independently.
When a model is updated on one target task, it is reverted to its state after training on the meta-training tasks before continuing with the next target task.
Metrics are aggregated using the mean across all out-of-fold tasks, and confidence intervals are computed using bootstrap resampling across tasks.

The same procedure is used when plotting performance with respect to training tasks, with the addition of randomly subsampling tasks within each training fold. There might be strategies for selecting an optimal diverse subset of tasks, we choose randomly for simplicity. The random subsample is without replacement, and seeded such that it is independent across folds but consistent amongst methods.

\subsection{Regret and Normalization}\label{app:regret_normalization}
As HPO and CASH are iterative processes, we measure regret as a curve over the number of configurations that have been evaluated.
As a summary statistic, regret is averaged over this curve.
Often, regret is reported as the difference between the optimal value known for an objective function and the optimal value among the configurations selected so far by a search strategy.
However, this does not take adaptive overfitting into account, where the selection of the \textit{best so far} may not translate to other test sets \citep{schroder_overfitting_2025}.
To prevent adaptive overfitting from positively skewing the results, we split validation and test folds.
The TabArena leaderboard data reports scores for individual folds per task; we set apart half of the folds to form the validation split and the other half as the test split.
The validation split defines $\mathcal{P}_{\mathrm{val}}$, which provides the matrix $P$, all target-task observations seen during search, and the selection of the \textit{best so far} configuration.
The held-out split defines $\mathcal{P}_{\mathrm{test}}$, which is used only for final reporting.
The final regret calculation therefore evaluates the test performance of the configuration selected by validation performance:

\begin{equation}\label{eq:test_regret}
    r_n(t) = \max_{\lambda \in \Lambda}\,\mathcal{P}_{\mathrm{test}}(t, \lambda)
    - \mathcal{P}_{\mathrm{test}}\!\left(
        t,\,
        \argmax_{\lambda \in \Lambda_{\mathrm{obs}}^{(n)}}\,\mathcal{P}_{\mathrm{val}}(t, \lambda)
    \right)
\end{equation}

\noindent where $n$ indexes the number of evaluated models, $\mathcal{P}_{\mathrm{test}}$ is performance on the test folds and $\mathcal{P}_{\mathrm{val}}$ on the validation folds.
In practice, we did observe some diverging of validation and test regret, shown in Section~\ref{app:adaptive_overfitting}).

Besides adaptive overfitting, we aim to eliminate another type of possible bias observed in related work: normalization leakage.
As some tasks may have consistently higher or lower performance across configurations, it is common to report normalized regret.
In the case of TabArena this is scaled such that the best score for a task is 1 and the median is 0.
This is justified when comparing AutoML methods from an outside perspective, but normalized values should not be available during the search process, as they contain information about the optimal achievable score.

To prevent leakage but keep performance on similar scales across tasks, we use two types of normalization.
During search, the performance matrix $P$ and observed target-task values are scaled by the dummy score.
The dummy score is the error achieved when constantly predicting the mean for regression tasks or the majority class for classification, which we consider a zero-cost value in practice that all methods would have access to.
When reporting normalized regret in graphs and tables, we use the aforementioned scaling that maps the best performance overall for a task to 1 and the median to 0.

Similarly, while the TabArena leaderboard imputes the few combinations where an algorithm is not capable of a specific task type with the score of a random forest, we score these combinations using the near zero-cost dummy score instead.

\subsection{Representation Encoding}\label{app:representation_encoding}
For the HPO benchmark, $\phi(\lambda)$ contains the hyperparameters of the single algorithm under consideration.
For the CASH benchmark, where $\lambda=(A,h)$, it concatenates the hyperparameter lists of all algorithms with a one-hot algorithm indicator.
Categorical choices are one-hot encoded and inactive or non-applicable hyperparameters are encoded consistently across all candidates in $\Lambda$.
For the vector-based performance variants, namely Linear-Perf, RF-Perf, GP-Perf, and MeanMMR, we use the full historical performance vector $P_\lambda$ directly and do not apply dimensionality reduction.
This differs from OBOE \citep{yang2019oboe}, which uses low-rank PCA-based representations; in early experiments, our method improved as the representation approached the full vector.

\subsection{Acquisition Rules}\label{app:acquisition_details}
Expected Improvement is used for surrogate models that provide a predictive mean and uncertainty estimate, while MMR-based acquisition is defined in the main paper.
For random forests, the predictive mean and standard deviation used by EI are estimated from the tree ensemble \citep{lindauer2022smac3}.
Given predictive mean $\mu(\lambda)$, standard deviation $\sigma(\lambda)$, and best observed performance $\mathcal{P}^*$, EI is:

\begin{align}
    \mathrm{EI}(\lambda) &=
    \left(\mu(\lambda)-\mathcal{P}^*\right)\Phi(Z)
    + \sigma(\lambda)\varphi(Z), \label{eq:ei}\\
    Z &= \frac{\mu(\lambda)-\mathcal{P}^*}{\sigma(\lambda)},
\end{align}

\noindent where $\Phi$ and $\varphi$ are the standard normal CDF and PDF, respectively.

For MeanMMR, the general MMR rule from the main paper is instantiated with the historical mean as the relevance score and cosine similarity on raw performance vectors as the diversity term:
\begin{equation}\label{eq:meanmmr}
    \mathrm{MeanMMR}(\lambda) =
    \alpha \cdot \frac{\sum\limits_{s\in T_{\mathrm{train}}}{P_{s, \lambda}}}{|T_{\mathrm{train}}|}
    - (1 - \alpha) \cdot \max_{\lambda' \in \Lambda_{\mathrm{obs}}}\, \cos(P_\lambda,\, P_{\lambda'})
\end{equation}
Because the first term does not depend on target-task observations, the historical mean ranking is fixed, and only the diversity term updates as configurations are evaluated on the target task.

\subsection{Gaussian Process Implementation}
\label{app:gp-details}
Gaussian processes (GP's) \citep{rasmussen_williams_2006} are the classical probabilistic surrogate in BO and provide a predictive posterior for EI.
We include a GP surrogate with both representations.
The Gaussian process surrogate is implemented using BoTorch's \texttt{SingleTaskGP} \citep{balandat2020botorch} with a Mat\'{e}rn-$\tfrac{5}{2}$ kernel \citep{rasmussen_williams_2006}.
Automatic Relevance Determination (ARD) is enabled, assigning an independent lengthscale to each input dimension so the model can identify which hyperparameters are most predictive.
Gamma priors are placed on the lengthscales ($\mathrm{Gamma}(3.0, 6.0)$) and output scale ($\mathrm{Gamma}(2.0, 0.15)$) to regularize the fit when few observations are available.
Inputs are min-max normalized to $[0,1]$ and outputs are standardised per task. Quantile scaling was attempted, which should automatically redistribute hyperparameter values that were originally sampled by TabArena authors at log-scale, but did not lead to an improvement.
Kernel hyperparameters are optimized by maximising the marginal log-likelihood via L-BFGS-B at each BO step, warm-started from the previous iteration's solution.

BoTorch's implementation is used across experiments and variants for consistency. Its performance relative to scikit-learn's \texttt{GaussianProcessRegressor} \citep{scikit-learn} with the default RBF kernel varies by benchmark: on the high-dimensional CASH benchmark, the simpler ARD-free scikit-learn kernel achieves lower regret for both representations (see Table~\ref{tab:main_results} in the main paper), whereas BoTorch performs better on the lower-dimensional HPO benchmark. Neither variant surpasses Linear-Perf.

\subsection{Random Forest Implementation}
\label{app:rf-details}
Both RF-Hyper and RF-Perf use scikit-learn's \texttt{RandomForestRegressor} \citep{scikit-learn} with \texttt{n\_estimators=100} and \texttt{random\_state=42}. RF-Hyper additionally sets \texttt{min\_samples\_leaf=3}, following SMAC's convention \citep{lindauer2022smac3} of preventing overly specific single-observation splits early in the search, when few configurations have been evaluated. RF-Perf uses scikit-learn's default \texttt{min\_samples\_leaf=1} and, like GP-Perf and Linear-Perf (Section~\ref{app:representation_encoding}), operates on the full performance vector without dimensionality reduction. Predictive mean and standard deviation for EI are estimated from the tree ensemble as described in Section~\ref{app:acquisition_details}.

\subsection{Summary of Final Hyperparameters}
\label{app:final_hyperparams}
For reproducibility, Table~\ref{tab:final_hyperparams} lists the final hyperparameter settings used for every surrogate model in the main experiments. Settings not listed follow scikit-learn or BoTorch defaults. Two-Tower settings were tuned per benchmark and are listed separately in Table~\ref{tab:two-tower-tuning}. Every model with a stochastic fitting procedure (Random Forest, GP restarts, Two-Tower initialization/training, and the meta-learning baselines) is seeded with \texttt{random\_state=42}; Bayesian ridge regression has no internal source of randomness.

\begin{table}[h]
\centering
\caption{Final hyperparameters used for every model in the main experiments. Exact configuration files are included in the code release.}
\label{tab:final_hyperparams}
\begingroup
\scriptsize
\setlength{\tabcolsep}{3pt}
\begin{tabularx}{\textwidth}{>{\raggedright\arraybackslash}p{2.0cm} >{\raggedright\arraybackslash}p{3.8cm} >{\raggedright\arraybackslash}X}
\toprule
\textbf{Model} & \textbf{Config key} & \textbf{Final hyperparameters} \\
\midrule
Linear-Hyper / Linear-Perf & \texttt{ridge\_ei} / \texttt{perf\_vector\_ridge\_ei} & BayesianRidge, scikit-learn defaults ($\alpha_1=\alpha_2=\lambda_1=\lambda_2=10^{-6}$, \texttt{max\_iter=300}, \texttt{tol=1e-3}); Linear-Perf fits without intercept (Section~\ref{app:bayesian_ridge}) \\
RF-Hyper & \texttt{rf\_ei} & \texttt{n\_estimators=100}, \texttt{min\_samples\_leaf=3}, \texttt{random\_state=42} \\
RF-Perf & \texttt{perf\_vector\_rf\_ei} & \texttt{n\_estimators=100}, \texttt{min\_samples\_leaf=1} (default), \texttt{random\_state=42} \\
GP-Hyper / GP-Perf & \texttt{botorch\_gp\_ei} / \texttt{botorch\_gp\_perf\_\{cash,hpo\}} & BoTorch \texttt{SingleTaskGP}, Mat\'{e}rn-5/2 ARD kernel, lengthscale prior $\mathrm{Gamma}(3.0,6.0)$, outputscale prior $\mathrm{Gamma}(2.0,0.15)$, L-BFGS-B MLL optimization (Section~\ref{app:gp-details}) \\
TwoTower-Hyper / TwoTower-Perf & \texttt{two\_tower\_hyper\_*} / \texttt{two\_tower\_perf\_*} & \texttt{random\_state=42}; remaining settings tuned per benchmark by random search, see Table~\ref{tab:two-tower-tuning} \\
\midrule
ABLR & \texttt{ablr} & feature dim 50, hidden layers $[50,50]$, 300 epochs, lr=0.01, \texttt{random\_state=42} \\
Active-Testing & \texttt{active\_testing} & similarity sharpness=1.0 \\
RGPE & \texttt{rgpe} & 256 bootstrap samples, 5 target CV folds, base-variance weighting enabled, weight-dilution prevention enabled, \texttt{random\_state=42} \\
RGPE-Oracle & \texttt{rgpe\_oracle} & as RGPE, with base-GP predictions replaced by true performance columns \\
TST-R & \texttt{tstr\_discrete} & bandwidth=0.5, base-variance weighting enabled, \texttt{random\_state=42} \\
MTGP & \texttt{mtgp\_\{hpo,cash\}} & 6 auxiliary tasks, 40 (HPO) / 48 (CASH) context flows, refit factor=3.0, \texttt{random\_state=42} \\
\bottomrule
\end{tabularx}
\endgroup
\end{table}

\subsection{Cost of GP-based Transfer and the RGPE-Oracle Ablation}
\label{app:gp-cost}
Results of the GP-based meta-learning baselines MTGP \citep{swersky2013multi} and RGPE \citep{feurer2018practical} are reported on the HPO benchmark only. Both fit exact Gaussian processes, whose cost grows cubically with the number of observations $n$. RGPE maintains one base GP for every prior task, scaling with $tn^3$ for $t$ tasks and $n$ configurations per task. MTGP includes all observations in one Gaussian process, thus scaling even faster following $(tn)^3$. In the experimental setup for HPO ($n=201$) this is affordable, on CASH ($n=3623$) it is not. For both cases, $t=40$ or $t=41$ depending on the fold.

To quantify this, we timed BoTorch's \texttt{SingleTaskGP} fit on the real CASH hyperparameter features. Single-run fit times were roughly $48$, $249$, and $1029$ seconds at $n=600$, $900$, and $1200$ respectively, that roughly matches the $O(n^3)$ cost and confirms that BoTorch does not fall back to a sub-cubic solver in this regime. Extrapolating to the full CASH set, a single base GP fit at $n=3623$ is on the order of six hours, so RGPE's roughly forty base GPs per fold amount to several days of pretraining per fold, before any target-task search. These are single measurements and intended only as orders of magnitude, but the gap to ABLR's $3.4$-minute pretraining, or Linear-Perf's none, is large enough that we exclude the GP-based methods from CASH rather than report downsampled results.

MTGP remains feasible on HPO only after subsampling: each fit conditions on a random subset of $40$ context configurations drawn from $6$ auxiliary tasks, without which the joint multi-task GP over context and target observations does not fit the HPO time budget. On CASH the fit would scale even more extremely than the several days of pretraining, given that it includes all tasks in one model.

To separate the cost of the GP fitting from the value of the ranking-transfer idea, we also evaluate RGPE-Oracle, a GP-free ablation that replaces each base-GP's predictions with the true performance column of the corresponding prior task. This removes the cubic cost and lets the method run on CASH. On HPO, RGPE-Oracle reaches an average normalized regret of $10.00$, competitive with the strongest methods and slightly ahead of the full GP-based RGPE ($10.57$), which indicates that the GP machinery adds little there. On CASH, where it can run but the full RGPE cannot, it reaches only $17.58$, well behind Linear-Perf ($12.19$). This supports the interpretation in the main paper's Discussion section that the non-negative ranking-agreement weighting over prior tasks is more restrictive than the signed linear combination fit by Linear-Perf.

\subsection{Two-Tower Implementation and Tuning}
\label{app:two-tower-tuning}
We use a Two-Tower surrogate as a collaborative-filtering-style meta-learning model. Let $z_t$ denote the embedding of task $t$ and $z_\lambda$ the embedding of configuration $\lambda$. The predicted performance of configuration $\lambda$ on task $t$ is given by their dot product, $\hat{\mathcal{P}}(t, \lambda) = z_\lambda^\top z_t$ \citep{huang2013learning,raza2026comprehensive}. It can learn across tasks when trained on the meta-training tasks, even when its configuration tower receives $x_\lambda^{\mathrm{hyper}}$. The two representation choices differ in how $z_\lambda$ is obtained: Two-Tower-Hyper maps $x_\lambda^{\mathrm{hyper}}$ to $z_\lambda$ through a learnable encoder, whereas Two-Tower-Perf uses no encoder and instead learns one free configuration embedding per column of the historical performance matrix by fitting $P_{t,\lambda}\approx z_t^\top z_\lambda$. This makes Two-Tower-Perf closely related to matrix factorization. Since the Two-Tower model does not produce a predictive posterior, EI is unavailable, and we instead use the MMR acquisition defined in the main paper with relevance score $\hat{\mathcal{P}}(t,\lambda)=z_\lambda^\top z_t$ and $z_\lambda$ the learned configuration embedding. The Two-Tower model offers a more nuanced comparison, being able to meta-learn using both hyperparameter or performance representations.

For Two-Tower-Hyper, the configuration tower is a multilayer perceptron that maps $x_\lambda^{\mathrm{hyper}}$ to $z_\lambda$.
We do not concatenate separate learned configuration latents to this input, so the variant remains purely hyperparameter-based.
The task tower is kept as an identity transformation from learned task latents $z_t$.
For Two-Tower-Perf, both towers are identity transformations over learned latents $z_\lambda$ and $z_t$, making it equivalent to a matrix-factorization-style model trained with the same procedure as Two-Tower-Hyper.
During target-task adaptation, only the task-side parameters are updated. MMR is used as acquisition rule instead of expected improvement due to the lack of posterior distribution over the output of the architecture.

For a fair comparison, it is valuable to also tune the top-level methods themselves.
We expected this to be most essential for the Two-Tower model, as a deep learning system is sensitive to settings such as learning rate and the dimensions of the multilayer perceptron.
Both Two-Tower variants were optimized through fifty iterations of random search on the first of the five folds, optimizing average normalized test regret.
This introduces a risk of overfitting and positively skewing the results for that model.
We consider this acceptable for the present comparison because the main intention is to compare variants within the same model family, both variants are treated the same, and the tuning fold accounts for only a fifth of the final score.
Alternatives such as nested cross-validation or a clean split with held-out tuning tasks would have required downscaling all experiments or accepting higher variance due to the smaller number of remaining tasks after holdout.

The shared search space covers three hyperparameters: embedding dimension sampled from $\{8, 16, 32\}$, learning rate sampled log-uniformly from $[0.005, 0.1]$, and batch size sampled from $\{256, 512, 1024, 2048\}$. Two-Tower-Hyper additionally searches over the hyperparameter-tower MLP architecture, with hidden-layer options $\{[16], [32], [64], [128], [16{,}16], [32{,}32], [64{,}64], [128{,}128]\}$, while Two-Tower-Perf has no MLP in the configuration tower (embeddings feed directly into the dot product). The optimal hyperparameters found per benchmark are listed in Table~\ref{tab:two-tower-tuning}.

\begin{table}[h]
\centering
\caption{Best hyperparameters found by random search for each Two-Tower variant and benchmark.}
\label{tab:two-tower-tuning}
\begingroup
\footnotesize
\begin{tabular}{llcccl}
\toprule
\textbf{Variant} & \textbf{Benchmark} & \textbf{Emb.\ dim} & \textbf{LR} & \textbf{Batch size} & \textbf{Hyper. tower} \\
\midrule
Two-Tower-Hyper & CASH & 8 & 0.007 & 1024 & $[64, 64]$ \\
Two-Tower-Hyper & HPO  & 8 & 0.007 &  256 & $[16]$ \\
Two-Tower-Perf  & CASH & 8 & 0.005 & 1024 & --- \\
Two-Tower-Perf  & HPO  & 8 & 0.020 &  512 & --- \\
\bottomrule
\end{tabular}
\endgroup
\end{table}

\begin{figure}[htbp]
  \centering
  \includegraphics[width=0.49\linewidth]{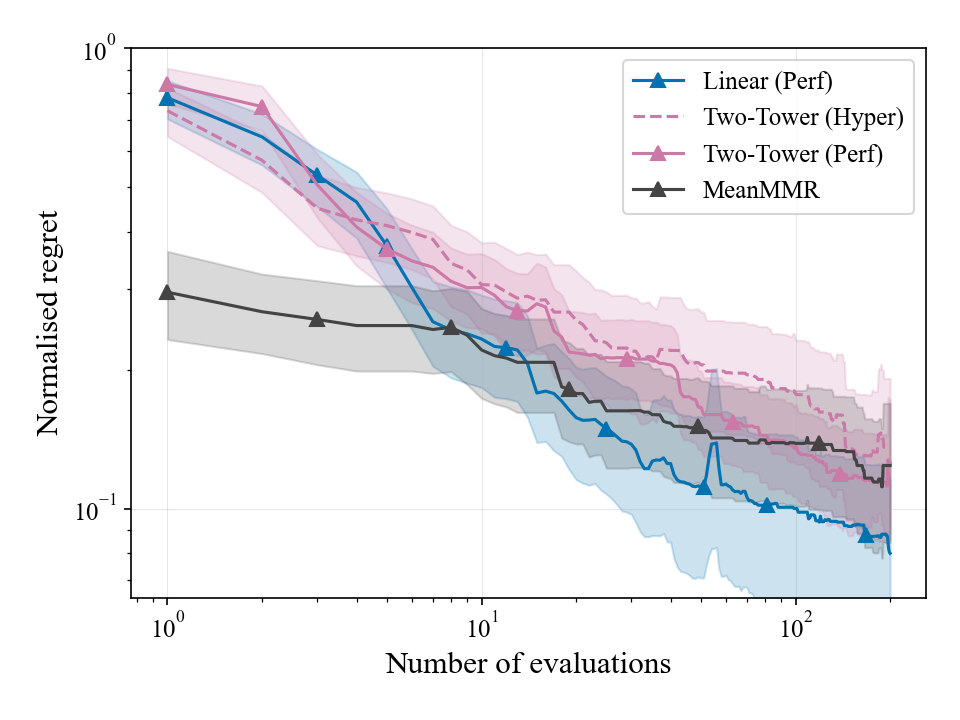}\hfill
  \includegraphics[width=0.49\linewidth]{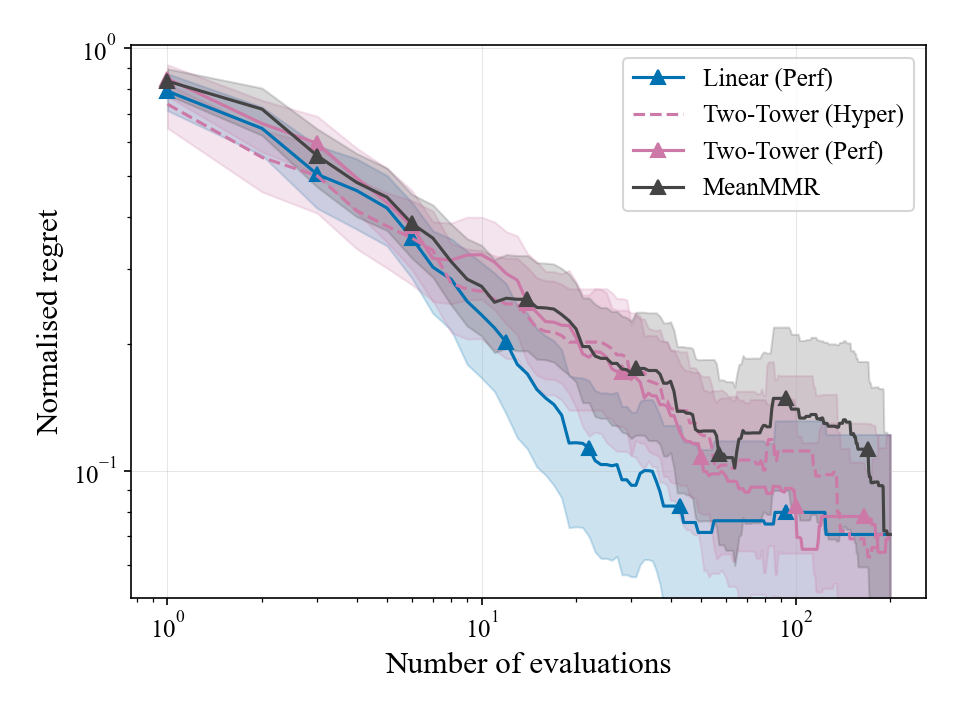}
  \caption{Test normalized regret of the Two-Tower variants on the CASH (left) and HPO (right) benchmarks, with MeanMMR and Linear-Perf as references. The two representations behave almost identically on HPO and similarly on CASH; neither clearly improves on MeanMMR, and both trail Linear-Perf. Shaded bands are 95\% bootstrap CIs across tasks.}
  \label{fig:two_tower_regret}
\end{figure}

Figure~\ref{fig:two_tower_regret} shows the test regret of both Two-Tower variants against MeanMMR and Linear-Perf on each benchmark. Performance on the HPO benchmark after tuning was almost the exact same between Two-Tower-Perf and Two-Tower-Hyper. This could suggest that training on the entire meta-dataset is sufficient for learning to map hyperparameter vectors to algorithm performance, providing evidence against claims that that mapping would be too complex of a task. However, it is also possible that Two-Tower-Perf does not see an improvement because of an unrelated bottleneck in the architecture that affects both variants. Neither variant clearly improves upon the baseline of MeanMMR, which should be reproducible in theory by a Two-Tower-Perf model with embedding size one where the configuration embedding value is the mean of the configuration. What that suggests is that while the model may be expressive enough, theoretical performance is lost as the training process does not find the optimal solution in practice. Increasing the number of epochs from 500 to 5000 did not yield a clear improvement. The Two-Tower variants were also observed to be highly sensitive to the training procedure parameters, unoptimized variants found during the tuning process sometimes performed worse than random search. So, while the Two-Tower model offers a good meta-learning solution for CASH in high-dimensional hyperparameter search space, the complexity of gradient-descent model training likely is the reason that it does not profit as much as other methods do when performance-based representations can be used directly.

\subsection{Compute}
\label{app:compute}
All experiments were run on a single machine using an AMD Ryzen 5 2600 Six-Core CPU, 16 GB of Memory and a GTX 1080 GPU, where only the Two-Tower model uses the GPU, running Windows with Python 3.11. Key library versions are scikit-learn 1.3.2, PyTorch 2.5.1 (CUDA 11.8), BoTorch 0.16.1, GPyTorch 1.15.2, and Hydra 1.3.2; exact pinned versions for all dependencies are given in the code release's \texttt{pyproject.toml} and \texttt{uv.lock}.
Table~\ref{tab:compute} reports total wall-clock time per model family across all tasks for both benchmarks including the scalability experiments, for 27h 20m in total.

\begin{table}[h]
\centering
\caption{Total wall-clock compute time per model family.}
\label{tab:compute}
\begin{tabular}{lr}
\toprule
\textbf{Model family} & \textbf{Wall-clock time} \\
\midrule
Two-Tower                 & 9h\,25m \\
Gaussian Process          & 7h\,43m \\
Random Forest             & 5h\,40m \\
Meta-learning baselines & 2h\,57m \\
Model-free             & 0h\,54m \\
Linear                    & 0h\,41m \\
\midrule
\textbf{Total}   & \textbf{27h\,20m} \\
\bottomrule
\end{tabular}
\end{table}

\section{Linear Surrogate Models}
\label{app:linear}
\subsection{Bayesian Ridge Formulation}\label{app:bayesian_ridge}
We use the Bayesian ridge regression formulation of \citet{mackay1992bayesian}, as implemented in scikit-learn's \texttt{BayesianRidge} estimator \citep{scikit-learn}.
Both Linear-Hyper and Linear-Perf use this estimator; for Linear-Perf, we disable the intercept to preserve the weighted-combination interpretation over prior tasks, which also gave a small empirical improvement.
For target-task observations $\mathcal{D}_t^{(n)}=\{(x_{\lambda_i}, y_i)\}_{i=1}^n$, the surrogate assumes a linear Gaussian likelihood,
\begin{equation}
    y_i = x_{\lambda_i}^{\top} w + \epsilon_i,
    \qquad
    \epsilon_i \sim \mathcal{N}(0,\alpha^{-1}),
\end{equation}
with an isotropic Gaussian prior over the regression weights,
\begin{equation}
    p(w\mid \eta)=\mathcal{N}(w\mid 0,\eta^{-1}I).
\end{equation}
Here $\alpha$ denotes the noise precision and $\eta$ the weight precision, corresponding to scikit-learn's coefficient-precision parameter \texttt{lambda\_}; we use $\eta$ to avoid overloading the configuration notation $\lambda$.
Both precisions are themselves assigned Gamma hyperpriors,
\begin{equation}
    \alpha \sim \mathrm{Gamma}(\alpha_1,\alpha_2),
    \qquad
    \eta \sim \mathrm{Gamma}(\lambda_1,\lambda_2),
\end{equation}
using the shape-rate parameterization, where we use the scikit-learn defaults $\alpha_1=\alpha_2=\lambda_1=\lambda_2=10^{-6}$.
Training is therefore not variational inference.
For fixed $\alpha$ and $\eta$, the posterior over $w$ is available in closed form as a Gaussian.
The precision parameters are then updated iteratively by empirical Bayes, also known as type-II maximum likelihood or evidence maximization, using the MacKay-style update rules described by \citet{mackay1992bayesian} and \citet{tipping2001sparse}.
We keep the default stopping settings, \texttt{max\_iter=300} and \texttt{tol=1e-3}.
Predictions use the resulting Gaussian posterior over $w$ to return both a predictive mean and standard deviation, which are passed to Expected Improvement.

\subsection{Choice of the Linear Surrogate}\label{app:linear_surrogate_choice}
A series of early experiments was done that resulted in the choice of Bayesian ridge regression. Initially, ordinary least squares optimization without regularization was attempted, referencing the existing work of OBOE \citep{yang2019oboe}. However, tuning the dimensionality reduction factor with PCA resulted in two findings. Firstly, predictive performance was most optimal when the rank equals the number of tasks (the full matrix rank). Secondly, there was a strong peak in prediction error whenever the number of observed target samples was equal to the rank of the dimensionality reduction. The peaking effect was most obvious when measuring prediction error per step, and not as easily seen when viewing regret curves over the entire process, possibly hiding this from previous authors. This peaking behavior is closely related to the classical peaking phenomenon and its modern discussion under double descent \citep{loog2020brief, viering2022shape}.
The instability of the unregularized model depending on the dimensionality motivated the use of Ridge regression without dimensionality reduction. This consistently provided better performance estimates without peaking behaviour.

Ridge regression was adapted to have an acquisition function either using non-probabilistic ridge regression with MMR, or Bayesian ridge regression using EI. When tuned well, both methods performed comparably. That also confirms the feasibility of using MMR as a replacement for EI. However, Ridge regression with MMR requires tuning the ridge regularization strength and MMR greediness, and Bayesian ridge regression with EI did not require any tuning, beyond always disabling the intercept. Hence Bayesian ridge regression was proposed and used to represent the linear model family.

Bayesian ridge regression re-estimates the noise and weight-prior precisions by
marginal-likelihood maximisation on every per-task fit. To test whether this
per-fit adaptation is the reason for any significant differences between the Linear-Perf model with respect to ABLR, we compare against a fixed-precision Bayesian
linear regression head identical to the one used by ABLR \citep{perrone_scalable_2018}. On both benchmarks the two are statistically
indistinguishable (average test normalized regret, CASH $12.19$ vs.\ $12.08$,
$p=0.76$; HPO $9.84$ vs.\ $9.96$, $p=0.20$; paired Wilcoxon signed-rank over
tasks). The adaptive prior precision estimation is therefore not the source of the
linear model's performance: the same fixed-precision head used by ABLR performs
equivalently when set once to a good value, supporting the use of the simpler Bayesian ridge surrogate that does not require setting the precision.

\subsection{Sign of the Learned Task Weights}\label{app:weight_signs}
\begin{figure}[htbp]
  \centering
  \includegraphics[width=0.6\linewidth]{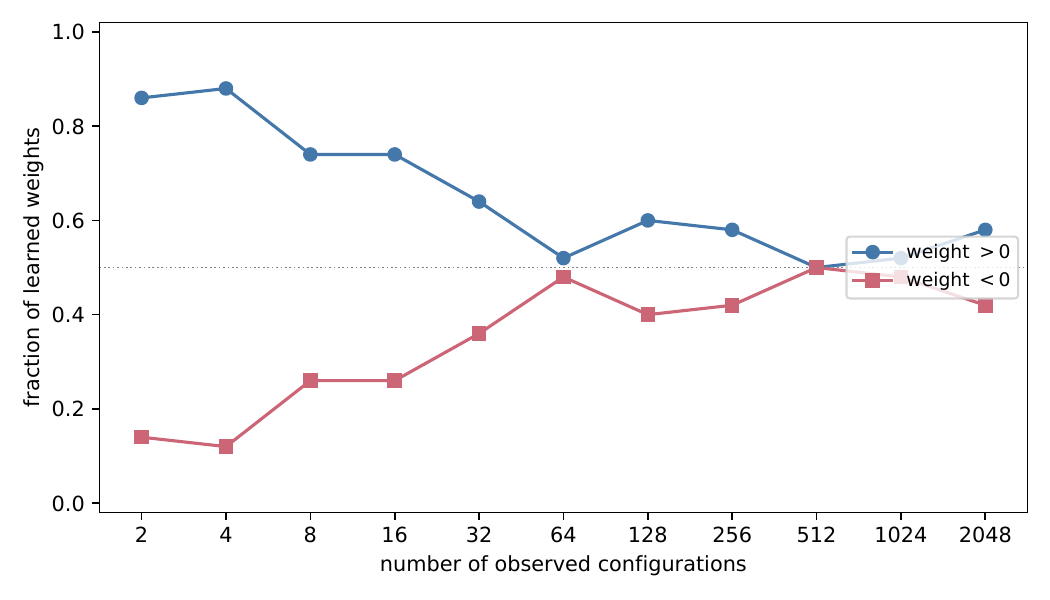}
  \caption{Sign of the learned Bayesian-ridge weights over prior tasks for a representative CASH target task using the performance representation: the fraction of weights that are positive versus negative as the number of observed target-task configurations grows.}
  \label{fig:bayesridge_weights}
\end{figure}

When Linear-Perf uses the historical performance vector as input, each regression coefficient weights performance on one specific prior task, so the fitted weights reveal how the target task is expressed as a combination of prior tasks. Figure~\ref{fig:bayesridge_weights} shows these weights for a representative CASH target task. As target-task observations accumulate, the fraction of negative weights grows to roughly half, far from the non-negative weights that a simple averaging of prior tasks would produce. The Bayesian ridge surrogate therefore does not merely average similar prior tasks; it also subtracts dissimilar ones. This signed combination is strictly more expressive than the non-negative ranking-agreement weights used by RGPE, consistent with RGPE-Oracle trailing Linear-Perf on CASH (see Section~\ref{app:gp-cost}).

\section{Adaptive Overfitting}
\label{app:adaptive_overfitting}
\begin{figure}[htbp]
  \centering
  \includegraphics[width=\linewidth]{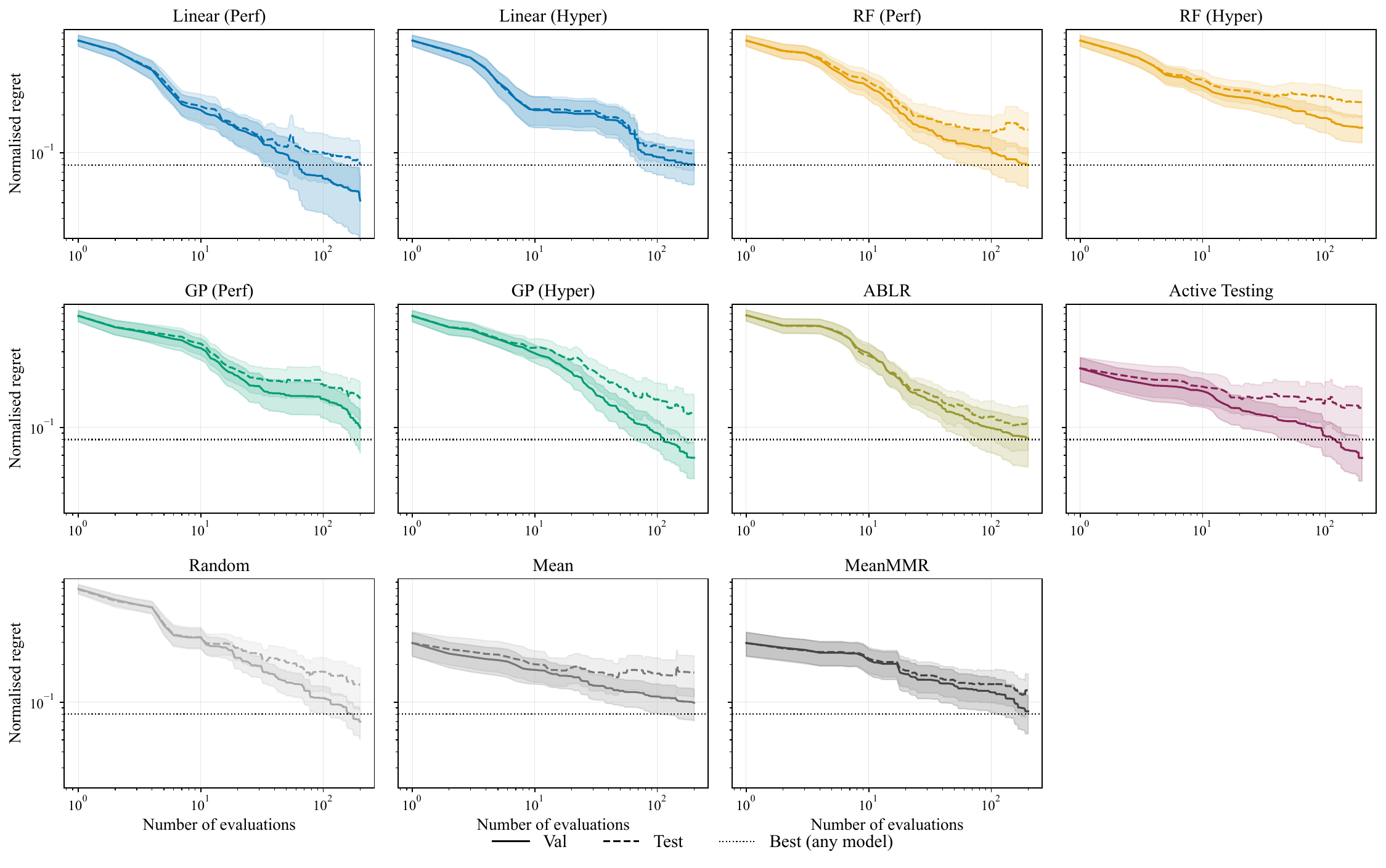}
  \caption{Validation and test regret for all methods on the CASH benchmark, with a horizontal line highlighting the best obtained test regret across methods within the 200 iterations.}
  \label{fig:adaptive_overfitting_cash}
\end{figure}

\begin{figure}[htbp]
  \centering
  \includegraphics[width=\linewidth]{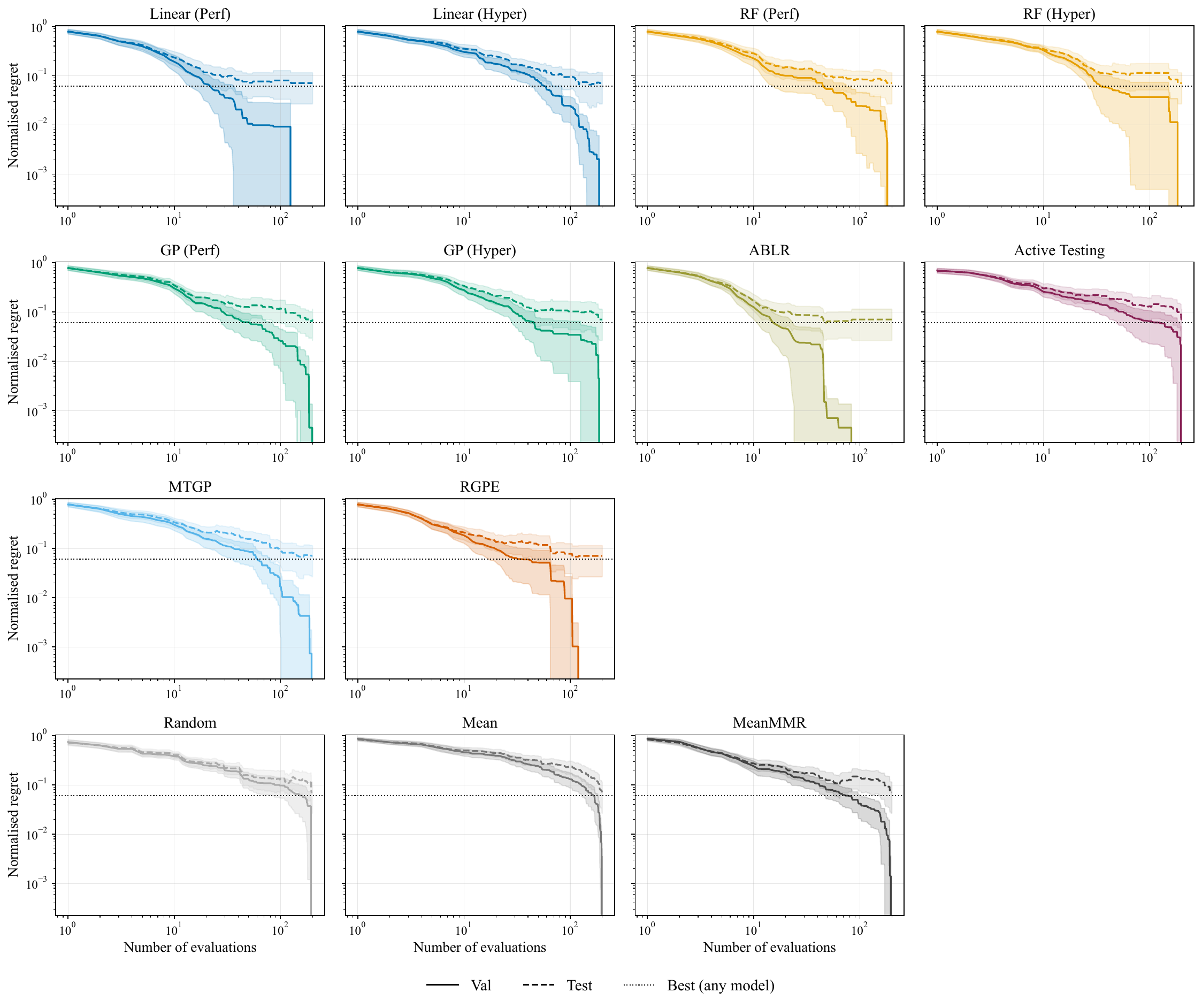}
  \caption{Validation and test regret for all methods on the HPO benchmark, with a horizontal line highlighting the best obtained test regret across methods within the 200 iterations.}
  \label{fig:adaptive_overfitting_hpo}
\end{figure}

As described in Section~\ref{app:regret_normalization}, we separate validation and test regret, and report only test regret in all main figures and tables.
Figure~\ref{fig:adaptive_overfitting_cash} and Figure~\ref{fig:adaptive_overfitting_hpo} compare the two metrics per method for the CASH and HPO benchmarks. Validation regret is consistently somewhat better than test regret, although this pattern is shared across methods and would not have significantly changed the main conclusions. The exception is a slightly larger gap for active testing. None of the methods appear to be overfitting to the extent that early stopping would have been necessary. We retain the validation-test split and reported test regret as it reflects more accurately how well a configuration might work when applied to real-world data after deployment that is not from the split it has been tuned on.

Across methods, validation and test regret track each other closely and decline together, with only minor gaps at higher evaluation counts (for example Active Testing on CASH). We therefore observe little adaptive overfitting in these results. On HPO the two curves stop improving simultaneously, which points to the limited search space, rather than overfitting to the validation split, as the factor that bounds further progress. We nonetheless retain the validation-test split as a conservative safeguard, since selecting the best-so-far configuration on the same data used to guide the search can in principle reward overfitting \citep{schroder_overfitting_2025}.

\section{TabArena Leaderboard Dataset}
\label{app:dataset}

\subsection{Base Model Set}
\label{app:base_models}

All models are taken from the TabArena benchmark leaderboard \citep{erickson2026tabarena}. For most models, the dataset contains results of one default configuration plus 200 randomly sampled configurations from the model's search space (201 total). Foundation models that were only evaluated at their default setting have a single configuration.

\begin{table}[h]
\centering
\caption{Base models used in the TabArena meta-dataset and the number of configurations evaluated per model.}
\label{tab:base_models}
\begingroup
\footnotesize
\begin{tabular}{llr}
\toprule
\textbf{Model} & \textbf{Category} & \textbf{Configurations} \\
\midrule
XGBoost              & Gradient boosting     & 201 \\
LightGBM             & Gradient boosting     & 201 \\
LightGBM (AIO)       & Gradient boosting     & 201 \\
CatBoost             & Gradient boosting     & 201 \\
Extra-Trees          & Tree ensemble         & 201 \\
Random Forest        & Tree ensemble         & 201 \\
k-NN                 & Classical             & 201 \\
Linear Model         & Classical             & 201 \\
MLP (Torch)          & Neural network        & 201 \\
MLP (FastAI)         & Neural network        & 201 \\
RealMLP              & Neural network        & 201 \\
ModernNCA            & Neural network        & 201 \\
TabM                 & Neural network        & 201 \\
xRFM                 & Neural network        & 201 \\
EBM                  & Interpretable         & 201 \\
TabPFNv2             & Foundation model      & 201 \\
RealTabPFN v2.5      & Foundation model      & 201 \\
TabDPT               & Foundation model      & 201 \\
TabICL               & Foundation model      &   1 \\
LimiX                & Foundation model      &   1 \\
Mitra                & Foundation model      &   1 \\
BetaTabPFN           & Foundation model      &   1 \\
TabFlex              & Foundation model      &   1 \\
\midrule
\textbf{Total}       &                       & \textbf{3623} \\
\bottomrule
\end{tabular}
\endgroup
\end{table}

\subsection{XGBoost Search Space}
\label{app:xgboost_search_space}

The XGBoost search space used by the TabArena authors to generate the 200 random configurations is shown below in Table~\ref{tab:xgboost_search_space}.

\begin{table}[h]
\centering
\caption{XGBoost hyperparameter search space. Log scale means the parameter is sampled uniformly in log space.}
\label{tab:xgboost_search_space}
\begingroup
\footnotesize
\setlength{\tabcolsep}{3pt}
\begin{tabular}{llll}
\toprule
\textbf{Hyperparameter} & \textbf{Type} & \textbf{Range} & \textbf{Scale} \\
\midrule
learning\_rate       & Float       & $[5 \times 10^{-3},\ 10^{-1}]$  & Log    \\
max\_depth           & Integer     & $[4,\ 10]$                        & Log    \\
min\_child\_weight   & Float       & $[10^{-3},\ 5.0]$                & Log    \\
subsample            & Float       & $[0.6,\ 1.0]$                     & Linear \\
colsample\_bylevel   & Float       & $[0.6,\ 1.0]$                     & Linear \\
colsample\_bynode    & Float       & $[0.6,\ 1.0]$                     & Linear \\
reg\_alpha           & Float       & $[10^{-4},\ 5.0]$                & Linear \\
reg\_lambda          & Float       & $[10^{-4},\ 5.0]$                & Linear \\
grow\_policy         & Categorical & \{depthwise, lossguide\}          & ---    \\
max\_cat\_to\_onehot & Integer     & $[8,\ 100]$                       & Log    \\
max\_leaves          & Integer     & $[8,\ 1024]$                      & Log    \\
\bottomrule
\end{tabular}
\endgroup
\end{table}

\end{document}